# Modeling and Validating Temporal Rules with Semantic Petri-Net for Digital Twins

Liu H.[a], Song X.[b], Gao G.[a,c,*], Zhang H.[a,c], Liu Y.[a,c], Gu M.[a,c]
[a] Tsinghua University, China. [b] Portland State University, United States. [c] Beijing National Research Center for Information Science and Technology (BNRist), China.
gaoge@tsinghua.edu.cn

**Abstract.** Semantic rule checking on RDFS/OWL data has been widely used in the construction industry. At present, semantic rule checking is mainly performed on static models. There are still challenges in integrating temporal models and semantic models for combined rule checking. In this paper, Semantic Petri-Net (SPN) is proposed as a novel temporal modeling and validating method, which implements the states and transitions of the Colored Petri-Net directly based on RDFS and SPARQL, and realizes two-way sharing of knowledge between domain semantic webs and temporal models in the runtime. Several cases are provided to demonstrate the possible applications in digital twins with concurrent state changes and dependencies.

## 1. Introduction

The digital twin technology integrates spatio-temporal data with domain semantic models, which enables tracking and simulating the change of building components, users and environment over time. There have been various applications of digital twins in construction and asset management stages in the construction industry.

Semantic rule checking on RDFS/OWL data have been widely used in the construction industry (Pauwels, et al., 2015; Beach, et al., 2015; Zhang, et al., 2019). At present, semantic rule checking is mainly performed on static models, in which the properties and relationships do not change in time. There have been researches about spatio-temporal rule checking, such as extending SPARQL for querying geometries (Battle and Kolas, 2011; Zhang, et al., 2018) and time periods (Koubarakis and Kyzirakos, 2010). However, digital twin applications usually have more complicated temporal states than just time periods, and also rule constraints to decide whether a state change is allowed. Such rules are closely related to semantic properties and relationships, so there is a requirement in integrated semantic and temporal rule checking.

Temporal modeling methods such as Finite State Machines (FSM) (Mealy, 1955; Moore, 1956), Colored Petri-Nets (CPN) (Jensen, 1987) and Business Process Modeling Notation (BPMN) (Decker, et al., 2010) are used in modeling the temporal states and rules. Related researches have focused on the interaction between temporal models and semantic webs. However, there are still challenges in two-way sharing of knowledge in the runtime for supporting integrated semantic and temporal rule checking.

In this paper, based on the CPN method, the Semantic Petri-Net (SPN) is proposed as a novel temporal modeling method. The temporal states and transitions are defined in RDFS, and the temporal transition rules are represented in SPARQL statements. As a result, SPN can run on semantic engines like Apache Jena and dotNetRDF, which enables direct usage of the vast semantic information in the domain semantic web such as ifcOWL (Beetz, et al., 2009) in the runtime of the temporal model. Experiments are performed to show the SPN application cases in modeling concurrent state change with dependencies and implementing automatic agents in the process maps.

## 2. Related Work

### 2.1 Temporal Modeling Methods

**FSM** is a popular method for modeling temporal states in the information industry. FSM represents a temporal model with a finite set of states, an input alphabet, and a global transition function defining the next state on each input. FSM is suitable for modeling systems in response to various signals or events, so that analysis tools such as temporal logic checking can be performed on the system.

**BPMN** is commonly used in engineering and project management, which represents events, activities and gateways in process maps. BPMN is suitable for modeling concurrent systems with multiple participants, such as a streamlined workflow. In the construction industry, BPMN is introduced in the Information Delivery Manual (IDM) (buildingSMART, 2010) method for representing the process maps of BIM data exchange.

**Petri-Net** and its variant methods are also widely used in modeling concurrent systems, among which the CPN is a method with both formality and ease of use. A Petri-Net is a graph with "places" to store tokens, "transitions" to consume and generate tokens, and "arcs" as the connection between places and transitions. CPN is a type of high-level Petri-Net with rules on places, transitions and arcs. Each token in a CPN can have an attached data value named a "color", which must be defined in a finite "colorset". A CPN can be equivalently unfolded to low-level Petri-Nets without colors, which enables mathematic tools (such as state space analysis and place invariant analysis) for applying temporal logic computation on the concurrent systems (Jensen and Kristensen, 2009).

CPN is powerful in modeling and analyzing both sequential and concurrent systems, and both FSM and BPMN can be implemented using CPN. It is also known that the extended CPN with an infinite colorset is Turing-complete (Peterson, 1980). In this paper, the Semantic Petri-Net is proposed based on the CPN, with the purpose to extend the ability in accessing the knowledge in semantic webs, and meanwhile to keep the downward compatibility to the mathematic tools for Petri-Nets.

### 2.2 Interaction between Temporal Models and Semantic Webs

Related researches have focused on the interaction between temporal models and semantic webs. The researches can be classified into the following topics.

**Describing the static structure of a temporal model using an ontology.** The contents of the ontologies are mainly about the temporal models themselves, but not much domain knowledge is included. The described temporal models include FSM (Belgueliel, et al., 2014), BPMN (Natschläger, 2011; Rospocher, et al., 2014; Annane, et al., 2019) and Petri-Nets (Gašević, 2004; Ma and Xu, 2009; Zhang, et al., 2011).

**Knowledge-based generation of temporal models.** Based on the knowledge graph of a certain domain, a temporal model is generated to perform as a runnable agent, for implementing a certain task such as a classifier (Yim, et al., 2011) or a workflow simulator (Wang, et al., 2007; Arena and Kiritsis, 2017). In such researches, domain knowledge is introduced in initializing the temporal models, but is no longer be referred to in the runtime.

**Knowledge sharing in the runtime of temporal models.** Such researches try to transfer information in the runtime either from the semantic web to the temporal model or in the opposite way. The Petri-Nets over Ontological Graphs (PNOG) (Szkoła and Pancerz, 2017) uses knowledge from the semantic web in the runtime of a Petri-Net, in which the tokens are with hierarchical classification, and the synonyms and hyponyms rules are supported. Knowledge-driven FSM (Moctezuma, et al., 2015) and DARPA Agent Markup Language (DAML) (Hendler, 2001) write the structure and the current state of an FSM into the semantic web, so that the current state and next states can be queried through a SPARQL endpoint.

Compared with the previous researches, the idea in this paper is to implement a runnable Petri-Net based on a semantic engine, for realizing the two-way sharing of knowledge between domain semantic webs and temporal models in the runtime, so that the SPARQL statements can be used inside the temporal model as transition rules concerning domain knowledge, and also outside the temporal model as domain queries concerning current temporal states.

A more detailed literature review of researches on knowledge sharing between temporal models and semantic models can be found in reference (Cheng and Ma, 2016).

## 3. Semantic Petri-Net

The definition of SPN is provided in section 3.1. The representation of SPN structure in RDFS is shown in section 3.2. The implementation of SPN rules in SPARQL is shown in section 3.3. The downward compatibility of SPN is discussed in section 3.4.

### 3.1 The Definition of SPN

For a semantic web $\mathbf{W}$, let $\mathbf{\Sigma}$ be its vocabulary, which is a finite set of terms (including literals and URIs) that are allowed in the semantic web. Using $\mathbf{\Sigma}$ as the colorset, the definition of SPN inherits from CPN, which is a tuple

$$\text{SPN} = \langle \mathbf{\Sigma}, \mathbf{P}, \mathbf{T}, \mathbf{A}, \mathbf{N}, \mathbf{C}, \mathbf{G}, \mathbf{E}, \mathbf{I} ; \mathbf{W} \rangle \tag{1}$$

$\mathbf{P}$, $\mathbf{T}$, $\mathbf{A}$, $\mathbf{N}$ are about the structure of an SPN, and $\mathbf{C}$, $\mathbf{G}$, $\mathbf{E}$, $\mathbf{I}$ are about the rules in the SPN. $\mathbf{P}$ is the set of places, $\mathbf{T}$ is the set of transitions, $\mathbf{A}$ is the set of arcs, and $\mathbf{N}$ is the assignment of each arc to link one place and one transition. $\mathbf{C}$ is the assignment of a subset of allowed colors to each place. $\mathbf{G}$ is the set of guard rules, which assigns each transition with a rule that returns a boolean value, deciding whether this transition is enabled. $\mathbf{E}$ is the set of arc expressions, which assigns each arc with an expression that returns a multiset of tokens, deciding which tokens to consume (place-to-transition arcs) or to generate (transition-to-place arcs). $\mathbf{I}$ is the assignment of an initial multiset of tokens to each place. Usually, the places are drawn as circles, the transitions are drawn as boxes, the arcs are drawn as directed arrows, and the tokens are drawn as little dots contained in the places.

The unit of the behavior of CPN is a "binding". A transition has a set of arguments, which is shared in the guard rules and the arc rules. A binding is an assignment of values to each argument, and a transition is enabled when the guard rule returns `true`, and there are enough tokens to consume in each input place according to the returned values of the corresponding arc expressions. SPN inherits the binding behavior of CPN, which is implemented by evaluating SPARQL queries and handling the tokens in the RDF graph.

### 3.2 SPN Structure Representation in RDFS

The RDFS classes and properties for representing SPN structures are with the namespace "`spn:`". The classes and properties for representing SPN structures are listed in Table 1.

**Transitions.** A transition is of the class `spn:Transition`, which may have a guard rule related with `spn:guardRule`. The arguments used in the guard rule and all related arcs are related with `spn:hasArg`. If the guard rule is not assigned, by default the transition is always enabled as long as enough input arguments are provided in a binding.

**Places.** Based on the Linked Data Platform (LDP) standard (W3C, 2015), the `spn:Place` is defined as a subclass of `ldp:Container`, in which the contained tokens are linked with `ldp:contains`. By implementing SPN place based on LDP container, the SPN place can be used in maintaining dynamic relationships in RDF, and the HTTP requests defined in the LDP standard can be adopted in manipulating the contained tokens in the SPN place. The `spn:colorRule` defines the allowed colors for the place, which can return a boolean value deciding whether a token is allowed. The `spn:initRule` defines the rule to get the initial tokens for the place.

**Arcs.** The `spn:Arc` is the abstract superclass for arcs, which must relate one place with `spn:relPlace` and one transition with `spn:relTransition`. The `spn:Arc` has two subclasses: `spn:ArcT2P` is for arcs pointing from a transition to a place, and `spn:ArcP2T` is for arcs pointing from a place to a transition. An arc must have at least one argument with `spn:hasArg`, and all arguments must also be assigned to the related transition. An arc expression can be related with `spn:arcExpr` to calculate the tokens to be consumed or generated. When there is only one argument, the arc expression can be null and the argument is directly consumed or generated.

Table 1: SPN structure components in RDFS.

| Class | Property | Range | Cardinality | Description |
|---|---|---|---|---|
| spn:Transition | spn:guardRule | spn:Rule | 0:1 | A boolean rule deciding whether the transition is enabled. |
| | spn:hasArg | spn:ArgDef | 1:? | Definition of arguments used in guard rules and arc rules. |
| spn:Place | ldp:contains | rdfs:Resource | 0:? | Contained tokens. |
| | spn:colorRule | spn:Rule | 0:1 | A boolean rule deciding whether a token is allowed in the place. |
| | spn:initRule | spn:Rule | 0:1 | A rule assigning initial tokens. |
| spn:Arc | spn:relPlace | spn:Place | 1:1 | Related place. |
| | spn:relTransition | spn:Transition | 1:1 | Related transition. |
| | spn:arcExpr | spn:Rule | 0:1 | An arc expression deciding which tokens to consume or to generate. |
| | spn:hasArg | spn:ArgDef | 1:? | Definition of arguments, must be a subset of the arguments of the related transition. |

### 3.3 SPN Rule Implementation in SPARQL

The SPN rules are formed as a tree structure in RDFS, in which each leaf node has a SPARQL statement. The classes and properties for representing SPN rules are listed in Table 2.

The `spn:Rule` is the abstract superclass of all SPN rule nodes, which has the following subclasses:

- The `spn:SPARQLRule` contains a SPARQL statement.
- The `spn:ConstantRule` contains a constant which can be any URI or literal value.
- The `spn:CompoundRule` contains multiple sub-rules connected by a logical operator

(`AND`, `OR`, `XOR`, or `NOT`), and each sub-rule returns a boolean value. Specifically, there can be only one sub-rule with a `NOT` operator.
- The `spn:ConditionRule` contains an "if" sub-rule returns a boolean value, and with the `true` and `false` conditions returned by the "if" sub-rule, the "then" and "else" sub-rules are checked respectively.

The rule nodes compose a rule tree structure, in which each leaf node should be an `spn:SPARQLRule` or an `spn:ConstantRule`. The SPARQL statements can be `ASK` queries returning a boolean value, or `SELECT` queries returning the queried tokens from the semantic web.

The `spn:ArgDef` is the class for defining the name and allowed types of an argument, which can be assigned to the transitions and arcs. The name of the argument is used as the input variable in SPARQL statements.

Table 2: SPN rule components in RDFS.

| Class | Property | Range | Cardinality | Description |
|---|---|---|---|---|
| `spn:SPARQLRule` | `spn:hasSPARQL` | `xsd:string` | 1:1 | The SPARQL statement of the rule |
| `spn:ConstantRule` | `spn:hasValue` | `rdfs:Resource` | 1:1 | The constant value of the rule |
| `spn:CompoundRule` | `spn:operator` | `xsd:string` | 1:1 | The logical operator. The allowed values are `AND`, `OR`, `XOR`, and `NOT`. |
|  | `spn:subRule` | `spn:Rule` | 1:? | The sub-rules connected by operator. |
| `spn:ConditionRule` | `spn:if` | `spn:Rule` | 1:1 | The "if" condition rule, which must return a boolean value. |
|  | `spn:then` | `spn:Rule` | 1:1 | The "then" condition rule when "if" condition returns `true`. |
|  | `spn:else` | `spn:Rule` | 1:1 | The "else" condition rule when "if" condition returns `false`. |
| `spn:ArgDef` | `spn:argName` | `xsd:string` | 1:1 | The argument name. |
|  | `spn:argType` | `rdfs:Class` | 0:? | Allowed argument types in a binding. |

### 3.4 Downward Compatibility of SPN

In CPN, the tokens are unrelated and the rules are static, so that the actions in a far-away transition do not change the behavior of a local transition, which is essential for the analysis of concurrent systems. While in SPN, the latent connections between tokens are introduced from domain semantic webs, so the local behavior may change due to a far-away transition. The benefit is that the structure of the Petri-Net can be simplified without explicitly representing the coupled states. However, the downward compatibility of SPN is necessary to ensure that the concurrency analysis methods for CPN are still applicable for SPN.

The conversion of the SPN to an equivalent CPN can be inspired by the "unfolding" of a CPN to lower-level Petri-Nets. Figure 1(a) shows an example unfolding of a CPN. A colored place is represented as multiple non-colored places corresponding to each allowed color of this place. A transition with conditions can be represented as multiple sub-transitions dealing with each condition, which can always be established by listing all possible bindings due to the "finite colorset".

In this section, a three-step unfolding process for SPN is proposed for obtaining an equivalent CPN, as shown in Figure 1(b).

**Step 1: putting all variables of the domain semantic web inside the SPN.** This ensures a "static domain semantic web" condition, in which the latent connections in the domain semantic web are not changeable by external editing operations. This can be done by representing all variable items in the domain semantic web as LDP containers, and representing the editing operations as transitions connected to the containers, since the LDP method is recommended for maintaining variable members and relationships in a semantic web.

**Step 2: splitting latent semantic connections into explicit arcs.** The SPARQL rules in SPN may concern latent connected remote tokens. In the "static domain semantic web" condition, for each SPARQL rule, a finite set of concerned remote tokens can be listed, and then a finite set of concerned remote places which may contain such tokens can be listed. As a result, the latent semantic connections can be split into explicit arcs. For each remote place, a pair of arcs are added for fetching the remote tokens and sending them back to keep the remote place unchanged.

**Step 3: implementing SPARQL rules in CPN.** Similar to the unfolding of CPN, due to the "finite vocabulary", the SPARQL rules can always be equivalently implemented in CPN, at least by listing all possible conditions of the bindings. There are also related researches about simplified unfolding methods (Liu, et al., 2012) rather than listing the bindings, which is out of the scope of this paper.

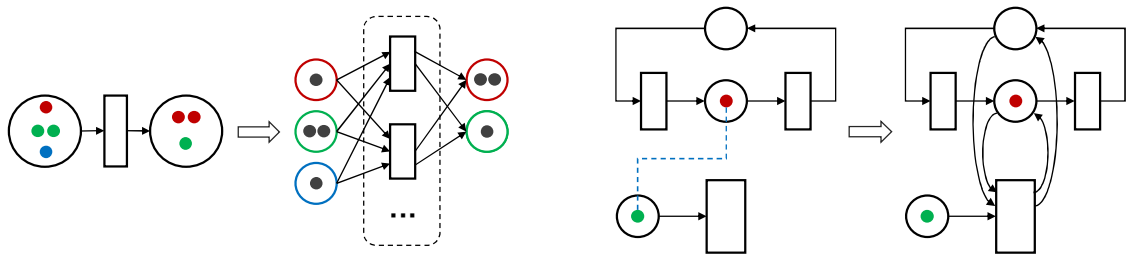

(a) unfolding a CPN to non-colored Petri-Net.   (b) unfolding an SPN with latent connections to CPN.

Figure 1: Examples of unfolding CPN and SPN.

## 4. Application Cases

In this section, two cases are provided to demonstrate the applications of SPN in modeling state changes with dependency in the construction industry: dependency checking in the construction process, and automatic agent for building information delivery. The system is developed based on the dotNetRDF as a SPARQL query engine. The experiments are performed on a PC with a 3.60GHz processor and 16GB of physical memory.

### 4.1  Case 1:  Dependency Checking in Construction Process

The digital twin for modeling the construction process of an asset can be represented as a spatio-temporal model in which each building object changes the state from "unbuilt" to "built". There are various dependency rules for deciding whether a state change is allowed, such as the amount of material and tools, the time period constraint, the state of host objects, or some permission files. In this section, an SPN use case is provided for modeling the construction process and checking the state change dependencies. A sample model of an office building (NIBS, 2012) is used in the experiment, as shown in Figure 2(a). The input data is a schedule of the starting and ending times for the construction phases of the levels, in which each phase corresponds to a discipline, including structure (ST), architecture (AR) and mechanical-electrical-plumbing (MEP), as shown in Figure 2(b).

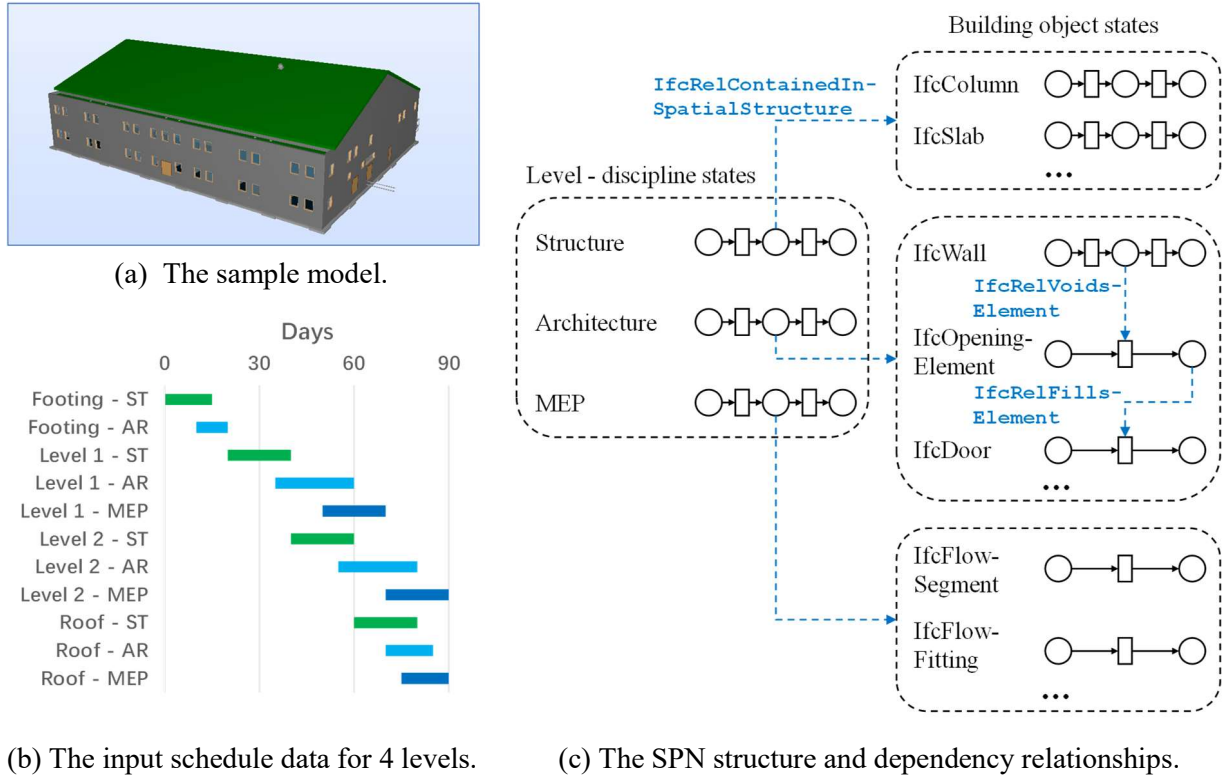

(a) The sample model.

(b) The input schedule data for 4 levels.   (c) The SPN structure and dependency relationships.

```
proj:T_Level_End_Struct  a                    spn:Transition;
                         bimspn:disciplineTag "STRUCT";
                         spn:guardRule        proj:cprule_0;
                         spn:hasArg           "?TOKEN".

proj:cprule_0   a             spn:CompoundRule;
                spn:operator  "NOT";
                spn:subRule   proj:sprule_1.

proj:sprule_1   a             spn:SPARQLRule;
                spn:hasSPARQL "ASK {
                    ?place a spn:Place.
                    ?place bimspn:disciplineTag ?dTag1.
                    ?SELF  bimspn:disciplineTag ?dTag2.
                    ?place bimspn:stateTag ?sTag.
                    FILTER (?dTag1 = ?dTag2 && ?sTag != 'END')
                    ?TOKEN ifc4:containsElements/ifc4:RelatedElements ?elem.
                    ?place ldp:contains ?elem. }".
```

(d) The RDF representation of an example transition and its guard rule nodes.

Figure 2: A use case of SPN in dependency checking in the construction process.

For each type of building objects, the states are represented as a series of places, and the dependencies are represented as the guard rules of the transitions. Small building components have two states (uninstalled and installed), and large components like walls and slabs have three states (not-started, in-processing and finished). The generated SPN structure is with 70 places and 43 transitions, as shown in Figure 2(c), with latent semantic connections as dashed arrows.

The states of the levels are concerned in the guard rules of other building objects with the "IfcRelContainedInSpatialStructure" relationships. The relationships between building objects are also included. For example, the openings must be installed during the construction of the host walls, and the installation of windows and doors must be after the finish of the host openings and walls. Other constraints added into the guard rules include maximum allowed tokens in each place, maximum triggering count allowed in one day for each transition, and minimum time span since the last state change for each token. In our experiment, there are in total 135 rules added to the SPN. Figure 2(d) shows the RDF representation of an example transition that ends the structure construction phase of levels, in which the guard rule requires that for each level (the "?TOKEN"), all structural objects contained in the level must be at the end place before the level is allowed to end the phase.

On initializing the model, the 4 levels and 7140 building objects from the IFC model are dispatched to the starting places as tokens. In our experiment, a clock ticks to simulate the days going by, and the guard rules are checked on each day to trigger the enabled transitions. The result shows whether the schedule is reasonable, or some sub-processes would stop with objects unfinished. In the experiment, the 90-days simulation finishes in 712 seconds with rules checked 441162 times and transitions triggered 8200 times in total.

Considering a real digital twin application, the transitions can be triggered manually or by sensors, and the guard rules are checked to ensure that the state changes are allowed, and to provide next-day recommendations according to a current state.

### 4.2 Case 2: Automatic Agent for Information Delivery Process

IDM is a standard method defining the process maps, use cases and information requirements in the delivery of BIM data between multiple parties. The SPN can be used to represent the process map, so that the states of the process can be queried through SPARQL. If automatic agents are included in the process map, the agent workflow can also be implemented in SPN.

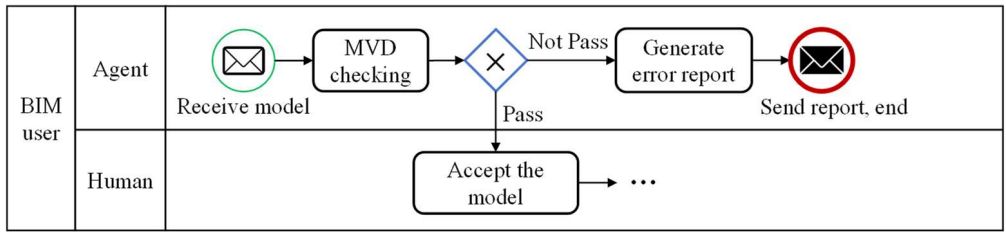

(a) A partial BPMN process map for BIM data exchange.

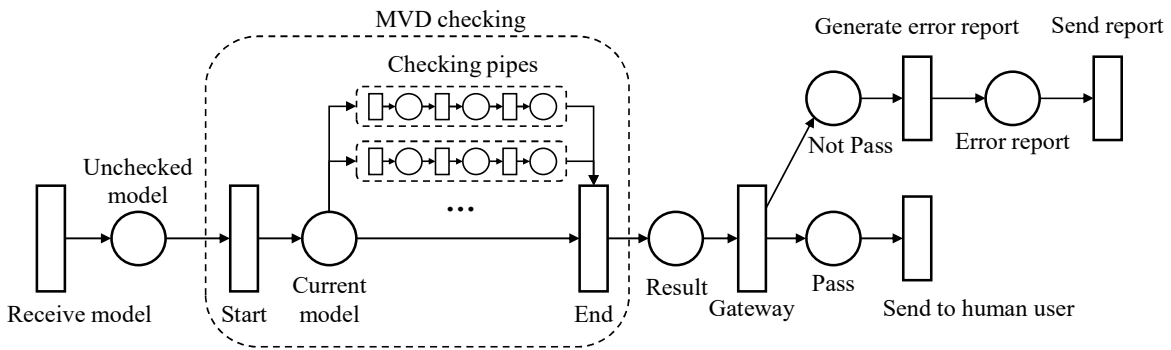

(b) The SPN implementation of the agent with MVD checking module.

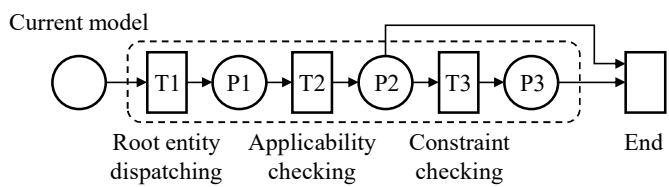

(c) The structure of an MVD checking pipe.

Figure 3: A use case of SPN in modeling the information delivery process.

The IDM method uses Model View Definition (MVD) for checking the compliance of data with information requirements. The MVD checking can be triggered by an automatic agent. Figure 3(a) shows a partial process map of BIM data exchange, in which an agent performs MVD checking on receiving the model to decide whether accept it. Figure 3(b) is the representation of the process map in SPN. Specifically, if the data is in RDF, the MVD checker can also be implemented with SPN. Figure 3(c) is the structure of a "checking pipe" for implementing the MVD checking process. An MVD rule is composed of an applicability rule and a constraint rule for a certain type of root entities. In running a checking pipe, first, the root entities are fetched from the model into place P1; next, the applicability rule is performed to filter the applicable entities into place P2; then, the constraint rule is performed for filtering the passed entities into place P3, and leaving unpassed entities in place P2.

In our experiment, the property existence rules from IFC4 Reference View MVD (buildingSMART, 2018) are used. An SPN is generated with 119 checking pipes and 1865 rule nodes. Using the same model in Figure 2(a) as input, the checking pipes are performed concurrently and the checking task finishes in 253 seconds.

## 5. Conclusion and Future Work

In this paper, the SPN is proposed as a novel temporal modeling and validating method directly based on RDFS and SPARQL, which realizes two-way sharing of knowledge between domain semantic webs and temporal models in the runtime. Compared with the current semantic model checking methods for BIM data, the SPN method explores a new scenario where the rule constraints are not only about the entities and relationships, but also about the process. Several cases demonstrate the ability of SPN in integrated rule checking involving both semantic information and temporal states, which shows the possible usage in digital twins with concurrent state changes and dependencies.

One topic for future work is to improve the usability of the current method. A more user-friendly way to compose the state model and define the constraints is needed for spreading the application in real projects. Another interesting topic is applying various concurrency analyses (such as state space analysis and place invariant analysis), and also involving logical quantifiers, especially those for temporal logic (such as Linear Temporal Logic and Computation Tree Logic) in SPN models, which can be helpful in complicated digital twin applications such as resource allocation and optimization.

**Acknowledgements**

This work was supported by the 2019 MIIT Industrial Internet Innovation and Development Project "BIM Software Industry Standardization and Public Service Platform".